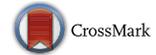

# Recent progress in semantic image segmentation

Xiaolong Liu[1] · Zhidong Deng[1] · Yuhan Yang[2]

**Abstract**
Semantic image segmentation, which becomes one of the key applications in image processing and computer vision domain, has been used in multiple domains such as medical area and intelligent transportation. Lots of benchmark datasets are released for researchers to verify their algorithms. Semantic segmentation has been studied for many years. Since the emergence of Deep Neural Network (DNN), segmentation has made a tremendous progress. In this paper, we divide semantic image segmentation methods into two categories: traditional and recent DNN method. Firstly, we briefly summarize the traditional method as well as datasets released for segmentation, then we comprehensively investigate recent methods based on DNN which are described in the eight aspects: fully convolutional network, upsample ways, FCN joint with CRF methods, dilated convolution approaches, progresses in backbone network, pyramid methods, Multi-level feature and multi-stage method, supervised, weakly-supervised and unsupervised methods. Finally, a conclusion in this area is drawn.

## 1 Introduction

Semantic image segmentation, also called pixel-level classification, is the task of clustering parts of image together which belong to the same object class (Thoma 2016).

Two other main image tasks are image level classification and detection. Classification means treating each image as an identical category. Detection refers to object localization and

This work was supported in part by the National Key Research and Development Program of China under Grant No. 2017YFB1302200 and by research fund of Tsinghua University - Tencent Joint Laboratory for Internet Innovation Technology.

✉ Zhidong Deng
   michael@mail.tsinghua.edu.cn

   Xiaolong Liu
   xllau@126.com

[1] State Key Laboratory of Intelligent Technology and Systems, Beijing National Research Center for Information Science and Technology, Department of Computer Science, Tsinghua University, Beijing 100084, China

[2] Department of Economics and Management, Tsinghua University, Beijing 100084, China

recognition. Image segmentation can be treated as pixel-level prediction because it classifies each pixel into its category. Moreover, there is a task named instance segmentation which joints detection and segmentation together. More details can refer to literature (Lin et al. 2014; Li et al. 2017a).

Semantic image segmentation has multiple applications, such as detecting road signs (Maldonado-Bascon et al. 2007), colon crypts segmentation (Cohen et al. 2015), land use and land cover classification (Huang et al. 2002). Also, it is widely used in medicine field, such as detecting brains and tumors (Moon et al. 2002), and detecting and tracking medical instruments in operations (Wei et al. 1997). Several applications of segmentation in medicine are listed in Dzung et al. (1999). In Advanced Driver Assistance Systems (ADAS) or self-driving car area, scene parsing is of great significance and it heavily relies on semantic image segmentation (Fritsch et al. 2013; Menze and Geiger 2015; Cordts et al. 2016).

Since the re-rising of DNN (Deep Neural Network), the segmentation accuracy has been significantly enhanced. In general, the methods before DNN are called traditional method. we also comply with this convention in the following sections. Traditional segmentation methods are briefly reviewed in this paper. More importantly, it will focus on the recent progress made by adopting DNN and organize them in several aspects. Moreover, we has carried out a survey on datasets of image segmentation and evaluation metrics.

This paper is organized as follows: Sect. 2 reviews the semantic image segmentation on datasets and evaluation metrics. Section 3 makes a brief summary of traditional methods. Section 4 introduces details of the recent progress. Finally, Sect. 5 makes a brief summary.

## 2 Datasets and evaluation metrics

This section reviews the the datasets related to semantic segmentation and evaluation metrics.

### 2.1 Datasets

At present, there are many general datasets related to image segmentation, such as, PASCAL VOC (Everingham et al. 2010), MS COCO (Lin et al. 2014), ADE20K (Zhou et al. 2017), especially in autonomous driving area Cityscapes (Cordts et al. 2016), and KITTI (Fritsch et al. 2013; Menze and Geiger 2015).

The PASCAL Visual Object Classes (VOC) Challenge (Everingham et al. 2010) consists of two components: (1) dataset of images and annotation made publicly available; (2) an annual workshop and competition. The main challenges have run each year since 2005. Until 2012, the challenge contains 20 classes. The train/val data has 11,530 images containing 27,450 ROI annotated objects and 6929 segmentations. In addition, the dataset has been widely used in image segmentations.

Microsoft COCO dataset (Lin et al. 2014) contains photos of 91 objects types which would be recognized easily by a 4-year-old person with a total of 2.5 million labeled instances in 328k images. They also present a detailed statistical analysis of the dataset in comparison to PASCAL (Everingham et al. 2010), ImageNet (Deng et al. 2009), and SUN (Xiao et al. 2010).

ADE20K (Zhou et al. 2017) is another scene parsing benchmark with 150 objects and stuff classes. Unlike other datasets, ADE20K includes object segmentation mask and parts segmentation mask. Also, there are a few images with segmentation showing parts of the heads (e.g. mouth, eyes, and nose). There are exactly 20,210 images in the training set, 2000 images in the validation set, and 3000 images in the testing set (Zhou et al. 2017). A group of images are shown in Fig. 1.

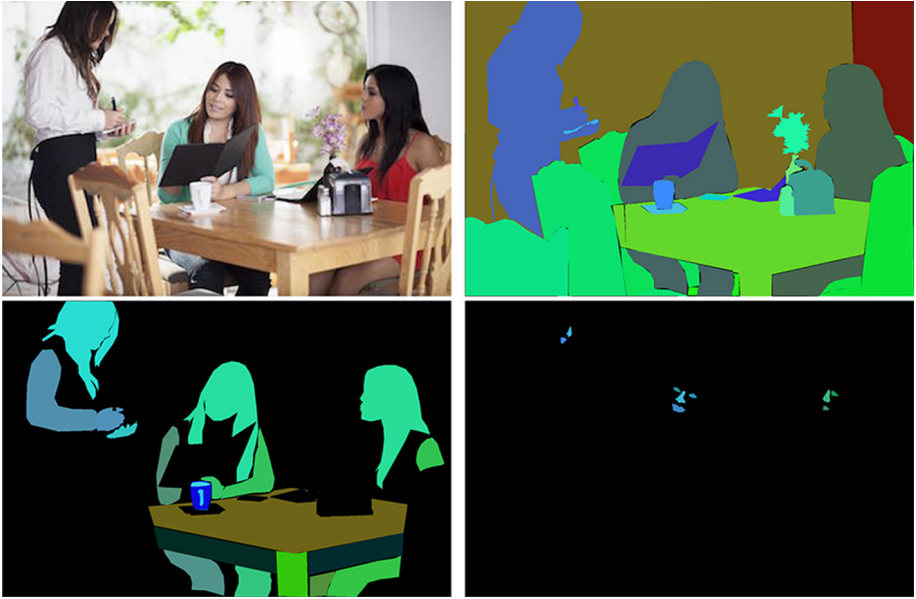

**Fig. 1** An example of ADE20K image. From left to right and top to bottom, the first segmentation shows the object masks. The second segmentation corresponds to the object parts (e.g. body parts, mug parts, table parts).The third segmentation shows parts of the heads (e.g. eyes, mouth, and nose)

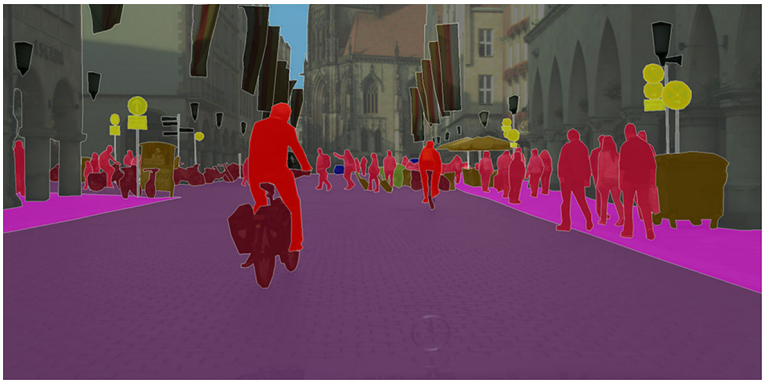

**Fig. 2** A fine annotated image from Cityscapes

The Cityscapes Dataset (Cordts et al. 2016) is a benchmark which focuses on semantic understanding of urban street scenes. It consists of 30 classes in 5000 fine annotated images that are collected from 50 cities. Besides, the collection time spans over several months, which covers season of spring, summer, and fall. A fine-annotated image is shown in Fig. 2.

KITTI dataset (Fritsch et al. 2013; Menze and Geiger 2015), as another dataset for autonomous driving, captured by driving around mid-size city of Karlsruhe, on highways, and in rural areas. Averagely, in every image, up to 15 cars and 30 pedestrians are visible.

The main tasks of this dataset are road detection, stereo reconstruction, optical flow, visual odometry, 3D object detection, and 3D tracking (http://www.cvlibs.net/datasets/kitti/).

In addition to the above datasets, there are also many others, such as SUN (Xiao et al. 2010), Shadow detection/Texture segmentation vision dataset (https://zenodo.org/record/59019#.WWHm3oSGNeM), Berkeley segmentation dataset (Martin and Fowlkes 2017), and LabelMe images database (Russell et al. 2008). More details about the dataset can refer to http://homepages.inf.ed.ac.uk/rbf/CVonline/Imagedbase.htm.

### 2.2 Evaluation metrics

Regular performance evaluation metrics for image segmentation and scene parsing include: pixel accuracy $P_{acc}$, mean accuracy $M_{acc}$, region intersection upon union (IU) $M_{IU}$, and frequency weighted IU $FW_{IU}$. Let $n_{ij}$ indicates the number of pixels of class i predicted correctly to belong to class j, where there are $n_{cl}$ different classes, and let $t_i = \sum_j n_{ij}$ indicates the number of pixels of class i. All of the four metrics are described as below (Long et al. 2014):

$$P_{acc} = \frac{\sum_i n_{ii}}{\sum_i t_i} \quad (1)$$

$$M_{acc} = \frac{1}{n_{cl}} \sum_i \frac{n_{ii}}{t_i} \quad (2)$$

$$M_{IU} = \frac{1}{n_{cl}} \sum_i \frac{n_{ii}}{t_i + \sum_j n_{ji} - n_{ii}} \quad (3)$$

$$FW_{IU} = \frac{1}{\sum_k t_k} \sum_i \frac{t_i n_{ii}}{t_i + \sum_j n_{ji} - n_{ii}} \quad (4)$$

## 3 Traditional methods

Before DNN is proposed, features and classification methods refer to the most important topics. In the computer vision and image processing area, feature is a piece of information which is relevant for solving the computational tasks. In general, this is the same sense as feature in machine learning and pattern recognition. Variety of features are used for semantic segmentation, such as Pixel color, Histogram of oriented gradients (HOG) (Dalal and Triggs 2005; Bourdev et al. 2010), Scale-invariant feature transform (SIFT) (Lowe 2004), Local Binary Pattern (LBP) (He and Wang 1990), SURF (Bay et al. 2008), Harris Corners (Derpanis 2004), Shi-Tomasi (Shi et al. 1994), Sub-pixel Corner (Medioni and Yasumoto 1987), SUSAN (Smith and Brady 1997), Features from Accelerated Segment Test (FAST) (Rosten and Drummond 2005), FAST- ER (Rosten et al. 2010), AGAST (Mair et al. 2010) and Multi-scale AGAST (Leutenegger et al. 2011) Detector, Bag-of-visual-words (BOV) (Csurka et al. 2004), Pselets (Brox et al. 2011), and Textons (Zhu et al. 2005), just to name a few.

Approaches in image semantic segmentation include unsupervised and supervised ones. To be specific, the simple one is thresholding methods which are widely used in gray images. Gray images are very common in medical area where the collection equipment is usually X-ray CT scanner or MRI (Magnetic Resonance Imaging) equipment (Zheng et al. 2010; Hu et al. 2001; Xu et al. 2010). Overall, thresholding methods are quite effective in this area.

K-means clustering refers to an unsupervised method for clustering. The k-means algorithm requires the number of clusters to be given beforehand. Initially, k centroids are randomly placed in the feature space. Furthermore, it assigns each data point to the nearest centroid, successively moves the centroid to the center of the cluster, and continues the process until the stopping criterion is reached (Hartigan and Hartigan 1975).

The segmentation problem can be treated as an energy model. It derives from compression-based method which is implemented in Mobahi et al. (2010).

Intuitively, edge is important information for segmentation. There are also many edge-based detection researches (Kimmel and Bruckstein 2003; Osher and Paragios 2003; Barghout 2014; Pedrycz et al. 2008; Barghout and Lee 2003; Lindeberg and Li 1997). Besides, edge-based approaches and region-growing methods (Nock and Nielsen 2004) are also other branches.

Support vector machine (SVMs): SVMs are well-studied binary classifiers which preform well on many tasks. The training data is represented as $(x_i, y_i)$ where $x_i$ is the feature vector and $y_i \in \{-1, 1\}$ the binary label for training example $i \in \{1, \ldots, m\}$. Where w is a weight vector and b is the bias factor. Solving SVM is an optimization problem described as Eq. 5.

$$\min_{w,b} = \frac{1}{2}||w||^2$$
$$s.t. \ \forall_{i=1}^{m} y_i \cdot (<w, x_i> + b) \geq 1 \quad (5)$$

Slack variables can solve linearly inseparable problems. Besides, kernel method is adopted to deal with inseparable tasks through mapping current dimensional features to higher dimension.

Markov Random Network (MRF) is a set of random variables having a Markov property described by an undirected graph. Also, it is an undirected graphical model. Let x be the input, and y be the output. MRF learns the distribution $P(y, x)$. In contrast to MRF, A CRF (Russell et al. 2009) is essentially a structured extension of logistic regression, and it models the conditional probabilities $P(Y|X)$. These two models and their variations are widely used and have reached the best performance in segmentation (http://host.robots.ox.ac.uk/pascal/VOC/voc2010/results/index.html; He et al. 2004; Shotton et al. 2006).

## 4 Recent DNN in segmentation

Artificial Neural Network (ANN) is inspired by biologic neurons. The basic element of ANN is artificial neuron. Each single artificial neuron has some inputs which are weighted and summed up. Followed by a transfer function or activation function, the neuron outputs a scale value. An example of neural model is illustrated in Fig. 3.

Based on artificial neuron, different stacking of the neurons forms Auto-encoder (Bengio 2009), Restricted Boltz- mann Machine (RBM) (Larochelle and Bengio 2008), Recurrent Neural Network or Recursive Neural Network (RNN), Convolutional Neural Network (CNN) (LeCun and Bengio 1995), Long Short Term Memory (LSTM) (Hochreiter and Schmidhuber 1997) and other types of ANNs. The basic architecture is illustrated in Fig. 4.

Convolutional Neural Network (CNN) (LeCun and Bengio 1995) uses shared-weight architecture, which is inspired by biological processes. The connectivity pattern between neurons is mimic of the organization of the animal visual cortex. Another important concept is receptive field, and it means that individual cortical neurons respond to stimuli only in a restricted region of the visual field. Also, they have the property of shift invariant or space invariant, based on their shared-weight architecture and translation invariance characteristics.

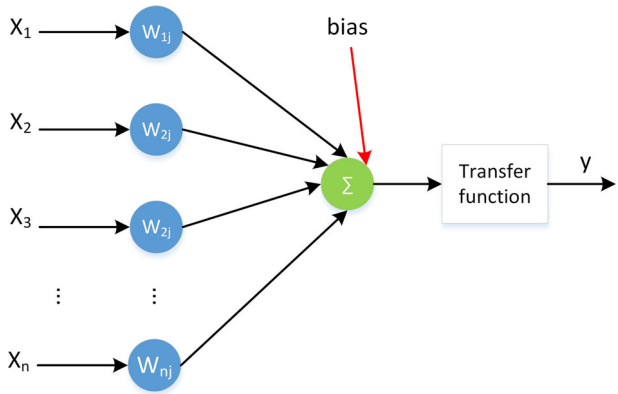

**Fig. 3** Artificial neuron model

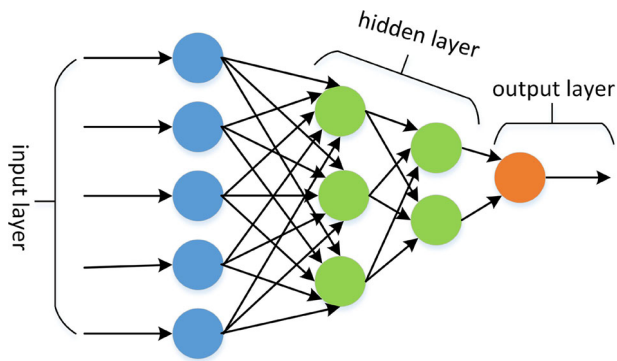

**Fig. 4** An example of artificial neural network model

Due to the excellent structure, CNN has obtained remarkable results on image classification, segmentation, and detection. The following part will present the recent progresses by applying CNNs in image semantic segmentation.

### 4.1 Fully convolutional network (FCN)

The paper (Long et al. 2014) is the first work that introduces ANNFCN to image segmentation area. The main insight is the replacement of fully connected layer by fully convolutional layer. With the use of the interpolation layer, it realizes that the size of output is the same as the input, which is essential in segmentation. To enhance the segmentation evidence, skips is adopted. More importantly, the network is trained end to end, takes arbitrary size, and produces correspondingly-sized output with efficient inference and learning.

FCN is implemented in VGG-Net and achieves the state of art on segmentation of PASCAL VOC (20% relative improvement to 62.2% mean IU in 2012) at that time, while the inference takes less than one fifth of a second for a typical image. The main architecture is shown in Fig. 5.

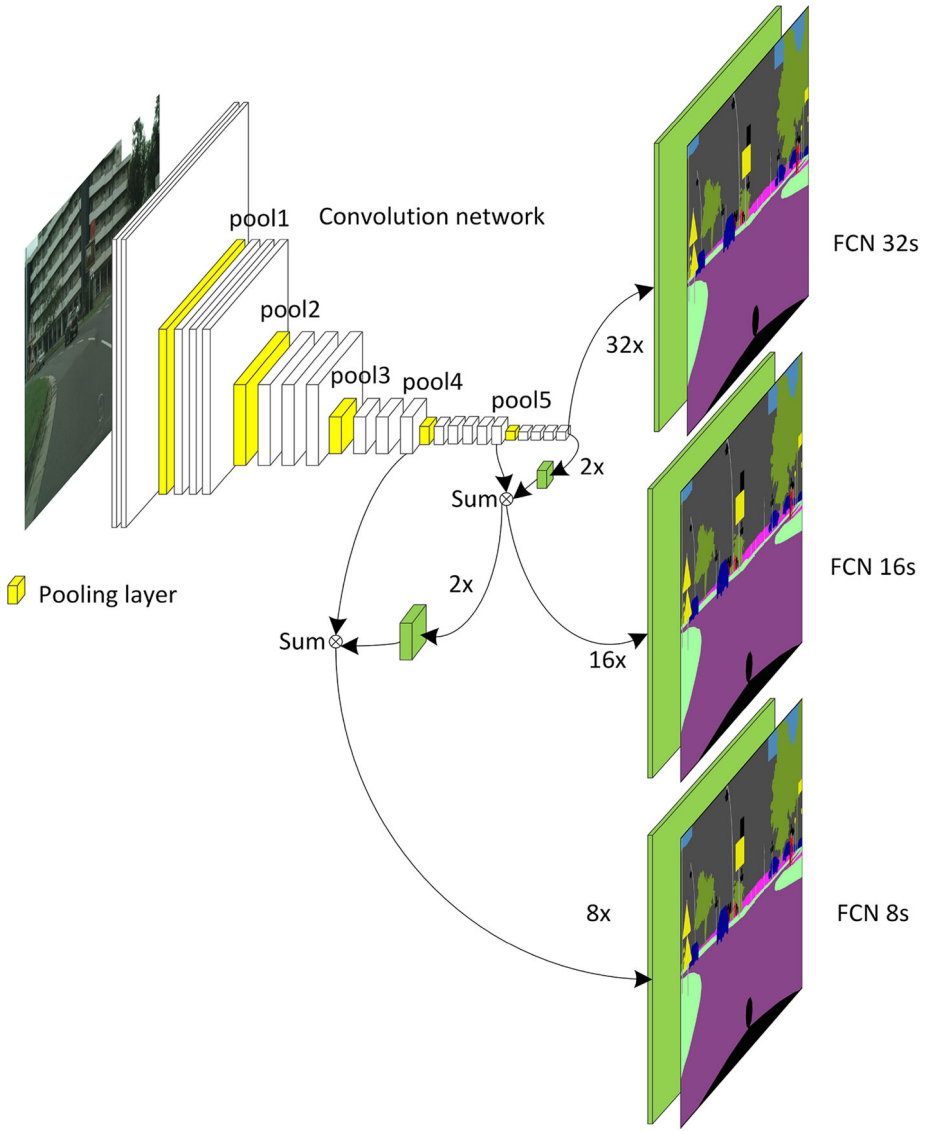

**Fig. 5** Fully convolutional network (FCN) architecture

## 4.2 Up-sample method: interpolation versus deconvolution

In addition to the FCN architecture, deconvolution layer is also adopted in semantic segmentation. The deconvolution network used in Noh et al. (2015) consists of deconvolution and un-pooling layers, which identify pixel-wise class labels and predict segmentation masks. Unlike FCN in paper (Noh et al. 2015), the network is applied to individual object proposals so as to obtain instance-wise segmentations combined for the final semantic segmentation.

Up-sample stage adopts bi-linear interpolation, which can refer to Long et al. (2014). Due to its computation efficiency and good recovery of the original image, the up-sample stage adopts bi-linear interpolation broadly. Deconvolution is the reverse calculation of convolution operation, which can also recover the input size. Thus, it can be applied into segmentation to recover the feature map size to original input size. The architecture implemented in Noh et al. (2015) is illustrated in Fig. 6. Also, other researchers implement semantic segmentation by deconvolution layer in different versions, which can refer to Mohan (2014), Monvel et al. (2003), Saito et al. (2016).

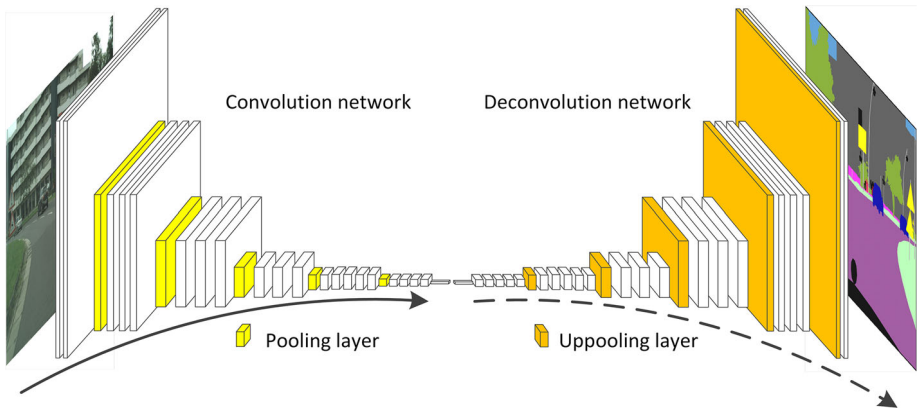

**Fig. 6** Deconvolution network architecture

### 4.3 FCN joint with CRF and other traditional methods

According to the research of Deeplab, the responses at the final layer of Deep Convolutional Neural Networks (DCNNs) are not sufficiently localized for accurate object segmentation (Chen et al. 2016b). They overcome this poor localization property by combining a fully connected Conditional Random Field (CRF) at the final DCNN layer. Their method reaches 71.6% IOU accuracy in the test set at the PASCAL VOC-2012 image semantic segmentation task. After this work, they carry out another segmentation architecture by combining domain transform (DT) with DCNN (Chen et al. 2016a) because dense CRF inference is computationally expensive. DT refers to a modern edge-preserving filtering method, in which the amount of smoothing is controlled by a reference edge map. Domain transform filtering is several times faster than dense CRF inference. Lastly, through experiments, it not only yields comparable semantic segmentation results but also accurately captures the object boundaries. Researchers also exploit segmentation by using super-pixels (Mostajabi et al. 2015; Sharma et al. 2015).

Paper (Liu et al. 2015) addresses image semantic segmentation by combining rich information into Markov Random Field (MRF), including mixture of label contexts and high-order relations (Figs. 7, 8, 9).

### 4.4 Dilated convolution

Most semantic segmentations are based on the adaptations of Convolutional Neural Networks (CNNs) that had originally been devised for image classification task. However, dense prediction, such as image semantic segmentation tasks, is structurally different from classification.

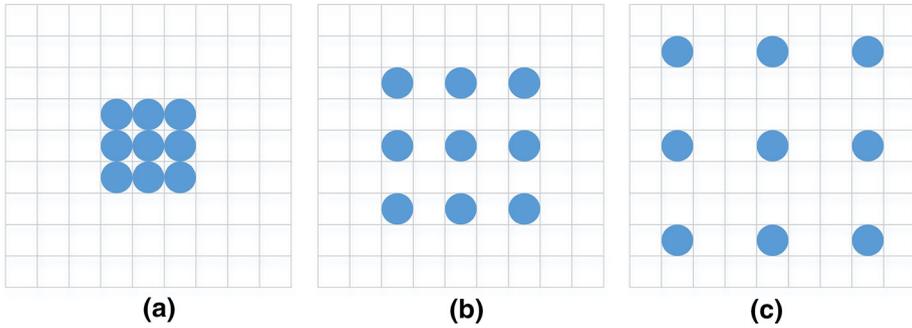

**Fig. 7** An example of dilated convolution (Atrous convolution or hole convolution). Convolution layer with kernel size 3 × 3, **a** normal convolution operation with parameter dilation = 1; **b** dilated convolution with parameter dilation = 2; **c** dilated convolution with parameter dilation = 3

Paper (Chen et al. 2016b) has already applied this strategy in their work. It is called 'Atrous Convolution' or 'Hole Convolution (Chen et al. 2016b)' or 'dilated convolution (Yu and Koltun 2015)'. Atrous convolution is originally developed for the efficient computation of the undecimated wavelet transform in the "algorithme à trous" scheme of paper (Holschneider et al. 1989). In Yu and Koltun (2015), they have presented a module using dilated convolutions to aggregate multi-scale contextual information systematically. The architecture is based on dilated convolutions that support exponential receptive field expansion without loss of resolution or coverage. Since the dilated convolution has griding artifacts, paper (Yu et al. 2017) develops an approach named dilated residual networks (DRN) to remove these artifacts and further increase the performance of the network.

### 4.5 Progress in backbone network

The backbone network refers to the main structure of the network. As is known to all, the backbone used in semantic segmentation is derived from image classification tasks. The FCN (Long et al. 2014) adopts VGG-16 net (Simonyan and Zisserman 2014) which did exceptionally well in ILSVRC14. Also, they consider AlexNet architecture (Krizhevsky et al. 2012) that won ILSVRC12 as well as GoogLeNet (Szegedy et al. 2015) that also did well in ILSVRC14. VGG net is adopted in many literatures, such as in Chen et al. (2016b) Liu et al. (2015).

After the release of ResNet (Deep residual network) (He et al. 2016) which Deeplab implement their work on which won the first place on the ILSVRC 2015 classification task, the semantic segmentation has made a new breakthrough. To find out the best configuration, paper (Wu et al. 2016a) evaluates different variations of a fully convolutional residual network, including the resolution of feature maps, the number of layers, and the size of field-of-view. Furthermore, paper (Wu et al. 2016b) studies the deep residual networks and explains some behaviors that have been observed experimentally. As a result, they derive a shallower architecture of residual network which significantly outperforms much deeper models on the ImageNet classification dataset.

Recently, ResNeXt (Xie et al. 2016) have been brought up as the next generation of ResNet. It is the foundation of our entry to the ILSVRC 2016 classification task in which we secured the 2nd place. GoogleNet also obtains development as Inception-v2, Inception-v3 (Szegedy et al. 2016), Inception-v4 and Inception-ResNet (Szegedy et al. 2017), which has already been adopted in the paper (Li et al. 2017b).

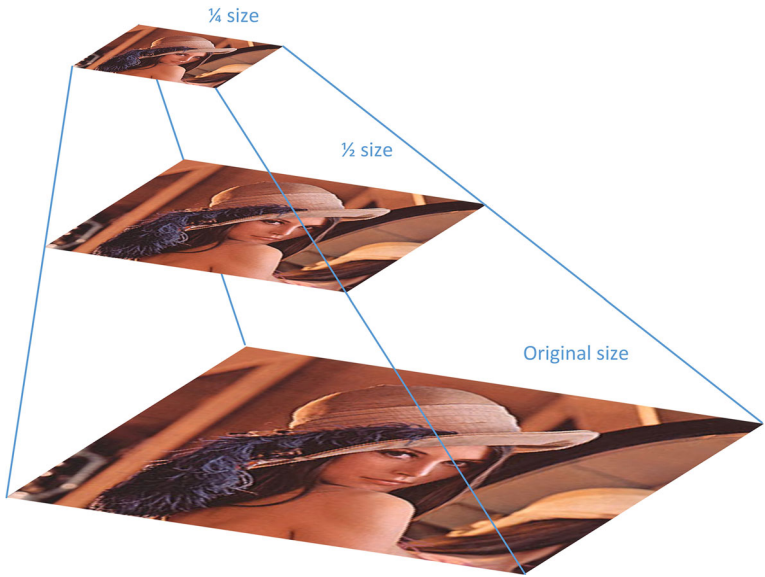

**Fig. 8** Three level image pyramid

### 4.6 Pyramid method in segmentation

Apart from adopting stronger backbone networks, researchers also attempt to combine pyramid strategy to CNN. The typical one is pyramid method.

1. Image pyramid

An image pyramid (Adelson et al. 1984) is a collection of images which are successively downsampled until some desired stopping criteria are reached. There are two common kinds of image pyramids: Gaussian pyramid which is used to downsample images and Laplacian pyramid which is used to reconstruct an upsampled image from an image lower in the pyramid (with less resolution).

In semantic image segmentation area, paper (Lin et al. 2016a) devises a network with traditional multi-scale image input and sliding pyramid pooling that can effectively improve the performance. This architecture captures the patch-background context. Similarly, Deeplab implements an image pyramid structure (Chen et al. 2016c) which extracts multi-scale features by feeding multiple resized input images to a shared deep network. At the end of each deep network, the resulting features are merged for pixel-wise classification.

Laplacian pyramid is also utilized in semantic image segmentation which can refer to paper (Ghiasi and Fowlkes 2016). They bring out a multi-resolution reconstruction architecture based on a Laplacian pyramid, which uses skip connections from higher-resolution feature maps and multiplicative gating to progressively refine boundaries reconstructed from lower-resolution feature maps. Paper (Farabet et al. 2013) presents a scene parsing system. The raw input image is transformed through a Laplacian pyramid. Meanwhile, each scale is fed to a two-stage CNN that produces a set of feature maps.

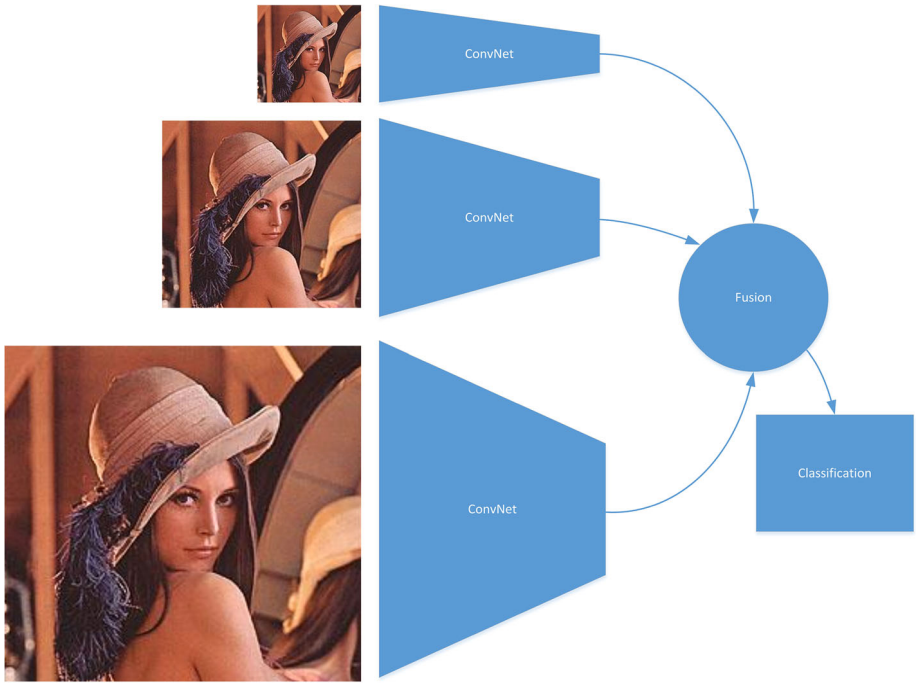

**Fig. 9** Image pyramid used in CNN

2. Atrous spatial pyramid pooling

Inspired by image pyramid strategy, (Chen et al. 2016b) proposes Atrous Spatial Pyramid Pooling (ASPP) to segment objects robustly at multiple scales. ASPP probes effective fields-of-views (FOV) and convolutional feature layer with filters at multiple sampling rates, and then captures objects image context at multiple scales. The architecture is shown in Fig. 10.

3. Pooling pyramid

Through pyramid pooling module illustrated in Fig. 11, paper (Zhao et al. 2016) exploits the capability of global context information by different-region based context aggregation and names their pyramid scene parsing network (PSPNet). Through experiments they report their outstanding results: with pyramid pooling, a single PSPNet yields new record of mIoU score as 85.4% on PASCAL VOC 2012 and 80.2% on Cityscapes.

The pyramid pooling adopts different scales of pooling size, then does up-sample process on the outputs to the original size, and finally concatenates the results to form a mixed feature representation. In Fig. 11, different scales of pooling sizes are marked with different colors. Generally speaking, the pyramid pooling can be applied to any feature map. For example, the application in Zhao et al. (2016) applies pyramid pooling in pool5 layer.

4. Feature pyramid

As pointed out by literature (Lin et al. 2016b), feature pyramid is a basic component in image tasks for detecting objects at different scales. In fact, recent deep learning object detectors have avoided pyramid representation because it is compute and memory intensive. In Lin et al. (2016b), they exploit the multi-scale, pyramidal hierarchy of CNN to construct feature pyramids with marginal extra cost. Also, Feature Pyramid Network (FPN) is developed for building high-level semantic feature maps at all scales.

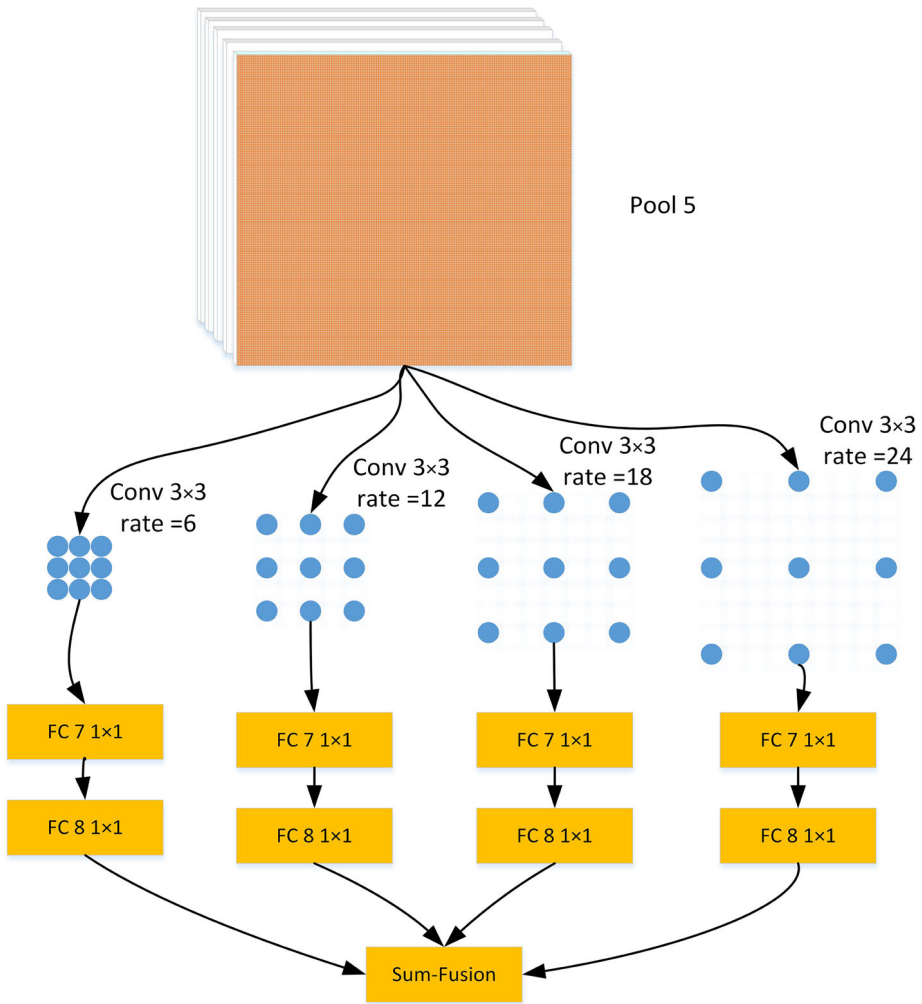

**Fig. 10** The atrous spatial pyramid pooling. (The distance in conv does not represent real rate)

### 4.7 Multi-level feature and multi-stage method

CNN can be treated as a feature extractor (Hariharan et al. 2015). Typically speaking, recognition algorithms based on convolutional networks (CNNs) use the output of the last layer as a feature representation. However, the information in this layer is too coarse for dense prediction. On the contrary, earlier layers may be precise in localization, but they will not capture semantics. To get the best of both advantages, they define the hypercolumns as the vector of activations of all CNN units above that pixel.

Indeed, skips have already been adopted in FCN (Long et al. 2014) which is depicted in Fig. 5. It seems that the multi-level method has been used in their work.

Multi-model is an ensemble way to deal with image tasks (Li et al. 2015; Viola and Jones 2001). Apart from multi-level strategy, a multi-stage method is used in semantic segmentation

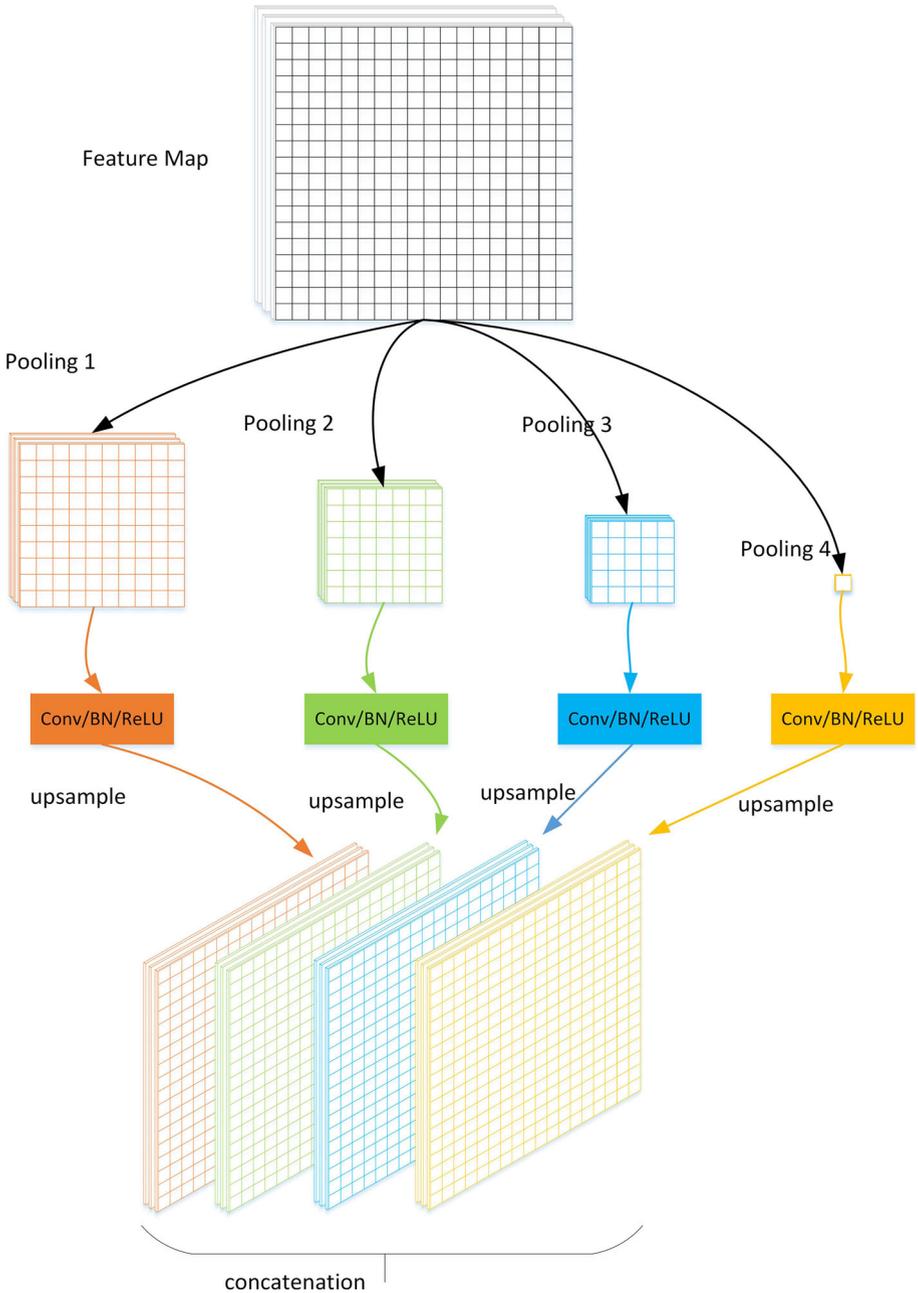

**Fig. 11** Illustration of pyramid pooling structure

(Li et al. 2017b). They propose deep layer cascade (LC) method to improve the accuracy and speed of semantic segmentation. Unlike the conventional model cascade (MC) (Li et al. 2015; Viola and Jones 2001) that consists of multiple independent models. LC treats a single deep model as a cascade of several sub-models and classifies most of the easy regions in the shallow stage and makes deeper stage focus on a few hard regions. It not only improves the segmentation performance but also accelerates both training and testing of deep network (Fig. 12).

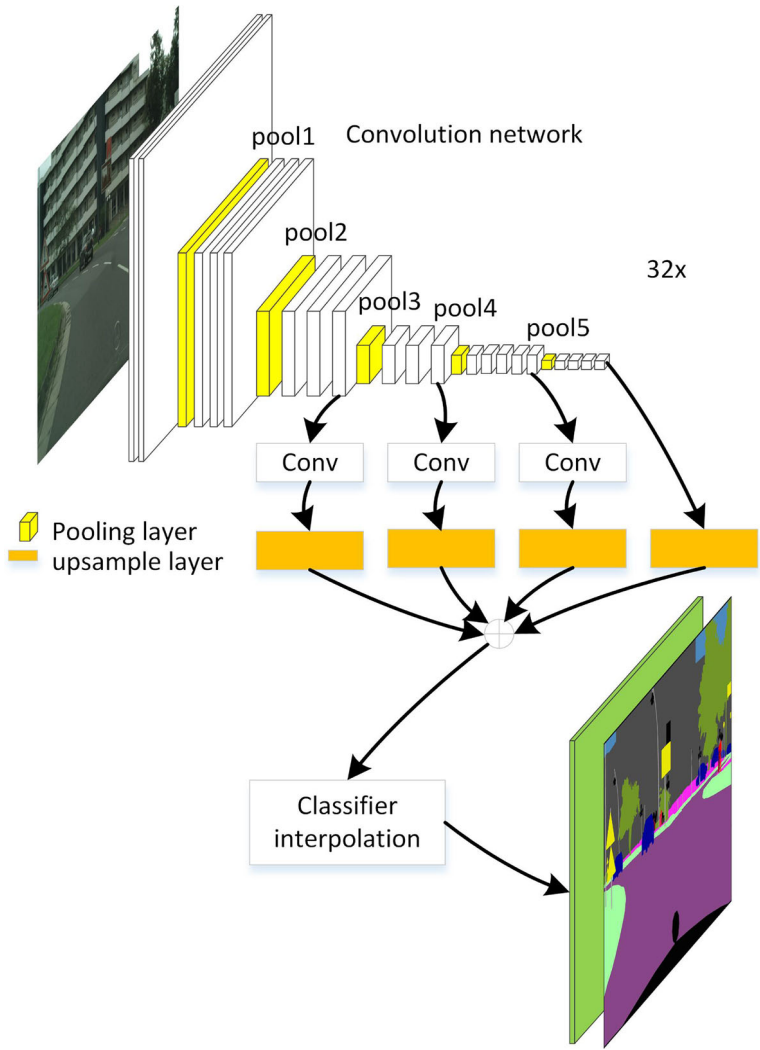

**Fig. 12** Structure adopted in Hariharan et al. (2015) as hypercolumns

### 4.8 Supervised, weakly-supervised and unsupervised methods

Most of the progress in semantic image segmentation are done under supervised scheme. However, researchers also dedicate to semi-supervised or non-supervised learning. More details can refer to Papandreou et al. (2015), Xia et al. (2013), Zhu et al. (2014), Xu et al. (2015).

## 5 Conclusion

Semantic image segmentation is a key application in image processing and computer vision domain. Besides briefly reviewing on traditional semantic image segmentation, this paper comprehensively lists recent progress in semantic image segmentation, especially based on DCNN, in the following aspects: 1. fully convolutional network, 2. up-sample ways, 3. FCN joint with CRF methods, 4. dilated convolution approaches, 5. progresses in backbone network, 6. pyramid methods, 7. Multi-level feature and multi-stage method, 8. supervised, weakly-supervised and unsupervised methods.

Till now, more and more methods are emerging to make semantic image segmentation more accurate or faster or both on accuracy and speed. We hope this review on recent progress of semantic image segmentation can make some help to researchers related to this area.